\pdfoutput=1

\documentclass[11pt]{article}

\usepackage[]{acl}

\usepackage{times}
\usepackage{latexsym}
\usepackage{tcolorbox}
\usepackage{subcaption}

\usepackage[T1]{fontenc}

\usepackage[utf8]{inputenc}

\usepackage{enumitem}
\usepackage{float}
\usepackage{graphicx}
\usepackage{multirow}
\usepackage{setspace}
\usepackage{hyperref}

\usepackage[linesnumbered, ruled]{algorithm2e}
\usepackage{dsfont}

\usepackage{mdframed}
\usepackage{booktabs}
\usepackage{graphicx}
\usepackage{subcaption}

\usepackage[normalem]{ulem}
\usepackage{enumitem}
\usepackage{xcolor}
\usepackage{soul}
\colorlet{soulred}{red!30}
\sethlcolor{soulred}%

\usepackage{amsmath,amsthm,amsfonts,amssymb,bm}
\usepackage{framed}
\usepackage{varioref}
\usepackage{float}
\usepackage{multirow}
\usepackage{dashrule}
\usepackage{url}
\usepackage{pifont}
\usepackage{helvet}

\usepackage{adjustbox} 

\definecolor{green}{rgb}{0.1,0.1,0.1}
\definecolor{chocolate}{HTML}{D2691E}
\definecolor{maroon}{HTML}{A00000}
\definecolor{indigo}{HTML}{4B0082}
\definecolor{green}{HTML}{008000}
\definecolor{cadmiumgreen}{rgb}{0.0, 0.42, 0.24}

\usepackage{amssymb}
\usepackage{pifont}

\makeatletter
\newcommand*\myfontsize{%
  \@setfontsize\myfontsize{8}{9}%
}
\makeatother

\makeatletter
\newcommand*\mysmallfontsize{%
  \@setfontsize\mysmallfontsize{7.4}{8.4}%
}
\makeatother

\usepackage{adjustbox}

\newcommand{\myskip}[1]{}

\newcommand{\ours}{\textsc{AALC}}

\title{\ours: Large Language Model Efficient Reasoning via \\ \underline{A}daptive \underline{A}ccuracy-\underline{L}ength \underline{C}ontrol}

\author{
\textbf{Ruosen Li}$^{1}$\textsuperscript{*}, \textbf{Ziming Luo}$^{1}$\textsuperscript{*}, \textbf{Quan Zhang}$^1$, \textbf{Ruochen Li}$^1$, \textbf{Ben Zhou}$^2$\\
\textbf{Ali Payani}$^3$, \textbf{Xinya Du}$^1$ \\
$^1$University of Texas at Dallas \quad
$^2$Arizona State University \quad
$^3$Cisco Research \\
\texttt{\{ruosen.li, ziming.luo, quan.zhang\}@utdallas.edu} \\
\texttt{\{ruochen.li, xinya.du\}@utdallas.edu} \\
\texttt{xzhou202@asu.edu} \quad \texttt{apayani@cisco.com}
}

\begin{document}
\maketitle
\renewcommand{\thefootnote}{}
\footnotetext{\textsuperscript{*}Both authors contributed equally to this work.}
\renewcommand{\thefootnote}{\arabic{footnote}}

\begin{abstract}
Large reasoning models (LRMs) achieve impressive reasoning capabilities by generating lengthy chain-of-thoughts, but this ``overthinking'' incurs high latency and cost without commensurate accuracy gains. 
In this work, we introduce \ours, a lightweight, accuracy-aware length reward integrated into reinforcement learning that dynamically balances correctness and brevity during training. By incorporating validation accuracy into the reward and employing a smooth, dynamically scheduled length penalty, \ours~delays length penalty until target performance is met. 
Through extensive experiments across standard and out-of-distribution math benchmarks, we show that our approach reduces response length by over 50\% while maintaining or even improving the original accuracy. 
Furthermore, qualitative analysis reveals that our method curbs redundant reasoning patterns such as excessive subgoal setting and verification, leading to structurally refined outputs rather than naive truncation. 
We also identify that efficiency gains are accompanied by reduced interpretability: models trained with \ours~omit some narrative framing and explanatory context. 
These findings highlight the potential of reward-based strategies to guide LRMs toward more efficient, generalizable reasoning paths.\footnote{Code is available at: \url{https://github.com/du-nlp-lab/AALC}

}
\end{abstract}

\section{Introduction}
Recent advances in Large reasoning models (LRMs)~\citep{openai2024learning, guo2025deepseek}, have significantly improved performance on complex reasoning tasks by scaling computational effort at inference time. This enables models to produce exploratory reasoning chains resembling internal deliberation, facilitating self-assessment and correction~\citep{openai2024learning, guo2025deepseek, qwen2024qwq}. Reinforcement learning (RL) has played a central role in enabling such capabilities. During RL training, models exhibit distinct reasoning phases characterized by longer, more intricate outputs and emergent strategies such as self-verification and decomposition~\citep{gandhi2025cognitive}. Empirical findings suggest a strong correlation between reasoning depth and accuracy gains~\citep{zeng2025simplerl, guo2025deepseek}.


Despite their benefits, extended reasoning chains often introduce inefficiencies due to unnecessary repetition and elaboration—a phenomenon known as overthinking~\citep{chen2024not}. This results in increased latency and computational cost with only marginal gains in accuracy. 
Motivated by Kimi1.5~\cite{team2025kimi}, which reports a reward function to shorten the chain-of-thought by using a length penalty, we proposed \ours\ reward to reduce models' inference length while maintaining models' performance.


This work presents two main contributions. We first introduce a lightweight reward mechanism that can be easily integrated into current reinforcement learning frameworks, featuring an accuracy-aware reward function that significantly reduces model reasoning length without compromising accuracy. Secondly, we provide an extensive evaluation across standard benchmarks to assess the trade-offs between response length and reasoning quality. Our method substantially reduces inference length while preserving both accuracy and semantic quality across standard and out-of-distribution math reasoning benchmarks, demonstrating the effectiveness and generality of our approach. In addition, our quantitative and qualitative analysis reveals:
\begin{enumerate}[itemsep=0pt,topsep=0pt,parsep=0.50pt,leftmargin=*]
    \item For each benchmark, we observe a critical inflection point in average response length: accuracy improves or remains stable above this point, but declines sharply below it. 
    \item Behavior analysis reveals that excessive ``subgoal setting'' and ``verification'' primarily drive response verbosity; our method effectively suppresses the frequency of these and two additional reasoning behaviors. 
    \item Models from the same base finetuned with \ours~\ tend to converge toward similarly concise reasoning lengths.
\end{enumerate}

\section{Related Work}

\paragraph{Efficient Reasoning}



Although increasing inference-time computational resources can improve the reasoning performance of LLMs, it often leads to overly verbose reasoning chains, resulting in higher computational costs and longer inference times~\citep{sui2025stop}. This “overthinking” phenomenon has motivated research into methods that encourage more concise and efficient reasoning.
To address this, several strategies have been proposed. Model-based approaches, such as reinforcement learning with length-aware rewards, train models to balance correctness with brevity (detailed in the next paragraph). Supervised fine-tuning on variable-length Chain-of-Thought (CoT) data, as used in Cot-valve~\citep{ma2025cot} and TokenSkip~\citep{xia2025tokenskip}, similarly promotes compact reasoning. Prompt-based methods like Token-Budget~\citep{han2024token} and Chain of Drafts~\citep{xu2025chain} guide models to be concise by explicitly constraining response length. Output-oriented techniques, including Coconut~\citep{hao2024training}, Softcot~\citep{xu2025softcot}, and Self-Calibration~\citep{huang2025efficient}, reduce token overhead by encoding reasoning steps in latent form. Finally, dynamic reasoning frameworks~\citep{liao2025reward, ding2025dynamic} adapt reasoning depth based on real-time feedback to optimize efficiency.

\paragraph{RL with Length Reward Design}
\label{sec:2.3}

Recent work has applied reinforcement learning (typically PPO~\citep{schulman2017proximal}) with explicit length-based rewards to control the length of Chain-of-Thought (CoT) reasoning. \citet{arora2025training} proposed a reward scheme favoring shorter correct answers, while \citet{yeo2025demystifying} introduced a Dirichlet-based Cosine Reward and length penalties to stabilize CoT growth. Kimi k1.5~\citep{team2025kimi} incorporates a length penalty into its policy optimization to improve long CoT activations and enable effective model merging. 
Some prior works either hurt models' performance while reducing the inference length or do not reduce inference length as much as we do~\cite{aggarwal2025l1, shen2025dast, arora2025training}.
Our method differs by incorporating validation accuracy directly into the reward function, enabling dynamic balancing between brevity and correctness.
Additionally, since our method only alters the reward computation, it remains compatible with various training algorithms, including PPO~\citep{schulman2017proximal} and GRPO~\citep{shao2024deepseekmath}.

\section{Methodology}
Current language models' outputs include much redundant content. However, a lightweight and effective length control mechanism is lacking to prevent overthinking.
In this section, we introduce a lightweight, length-controlled reward designed to prevent overthinking without compromising model performance.
Unlike most prior works that rely solely on the length of predictions and reference texts to regulate inference length, we propose a more principled reward design that incorporates validation accuracy to better balance performance and brevity. 
Section~\ref{sec:length_reward} introduces our length reward formulation, which follows the standard structure adopted in prior work. Section~\ref{sec:dynamic_attention} extends this by incorporating accuracy into the reward through a dynamic attention mechanism, effectively combining both objectives to prevent overthinking without sacrificing accuracy. Finally, Section~\ref{sec:scheduling} presents two dynamic scheduling strategies tailored for GRPO~\citep{shao2024deepseekmath} training, allowing the reward signal to adapt for training. 


\subsection{Length Reward}
\label{sec:length_reward}

To maintain model performance, it is crucial to incorporate accuracy into the design of the reward function. Specifically, the length reward is activated only when the validation accuracy meets a particular condition; otherwise, the model is encouraged to focus on improving accuracy before optimizing length.
At the step $i$ of the GRPO training process, the following equations define the length reward for a prediction:
\begin{align*}
    r_\text{acc} &= \frac{A_{\text{val}}}{A_\text{target}} \\
    r_\text{len} &= \min(1, \frac{L_\text{pred}}{L_{\text{max}}}) \in [0,1] \\
    R_\text{len} &= 1 - \min(r_\text{acc}^{\beta}, r_\text{len}) \in [0 ,1]
\end{align*}
where $A_\text{val}$ and $A_\text{target}$ denote the validation accuracy and the dynamically scheduled target accuracy at iteration $i$
$L_\text{pred}$ is the length of the model's predicted output, and $L_{\text{max}}$ is the preset maximum allowed length;
$r_\text{acc}$ and $r_\text{len}$ represent the normalized accuracy and length, respectively.
$R_\text{len}$ is the length reward. A value close to 1 implies that the rear is inactive (no penalty), either because the output is sufficiently short or the model has not yet achieved enough accuracy to trigger length constraints.
The hyperparameter $\beta$ controls the sensitivity of the reward to accuracy. A larger $\beta$ delays the activation of the length penalty until higher accuracy is achieved, while a smaller $\beta$ allows earlier enforcement of length constraints.
Instead of applying a hard threshold on accuracy, these formulations use a smooth transition to modulate the length reward, ensuring continuous control that adapts dynamically as the model improves.

\subsection{Dynamic Attention to Accuracy and \ours}
\label{sec:dynamic_attention}



During training, accuracy remains the primary objective, particularly in the initial stage, which we refer to as the "warm-up" phase when the model is still learning to produce correct outputs. As validation accuracy reaches its peak, the focus gradually shifts toward reducing generation length while maintaining accuracy. To automatically handle this transition, we introduce a dynamic attention mechanism controlled by validation accuracy. Additionally, we combine the focus on accuracy and length through an interpolation strategy.

The raw reward only focuses on correctness but overlooks text length. It is defined as $R_\text{raw}=\mathbb{I}(y_\text{pred}, y_\text{gold})$, where $R_\text{raw}=1$ indicates that the prediction contains the correct answer and 0 otherwise.
To balance performance and brevity, we introduce \textit{Accuracy Attention}, a mechanism that dynamically adjusts the weight of the raw reward based on the model’s current accuracy.
The accuracy attention $Att_\text{acc}$ is computed as:
\begin{align*}
    Att_\text{acc} = \gamma + (1 - \gamma) (1 - r_\text{acc})
\end{align*}
where
$\gamma$ is a hyperparameter to control the minimum attention to accuracy.
Initially, the relatively low validation accuracy emphasizes the raw reward to boost performance. As accuracy improves, the focus shifts to the length reward, encouraging more concise predictions.
The \ours~ reward is the weighted combination of the raw reward $R_\text{raw}$ and the length reward $R_\text{len}$.
\begin{align*}
    \text{AALC} = Att_\text{acc} \cdot R_\text{raw} + \alpha \cdot R_\text{len}
\end{align*}
where $\alpha$ controls the impact of the length reward. 
A large $\alpha$ will hurt models' performance. It will be discussed in section \ref{sec:ablation-study}.
Notably, the reward function serves two goals:
(1) to guide the model toward accurate predictions when performance is suboptimal, and
(2) to promote concise reasoning when accuracy is sufficiently high.

\begin{table*}[ht]
\centering
\begin{tabular}{lrrrrrr}
\toprule
                                     & \multicolumn{3}{c}{\textbf{GSM8k}}           & \multicolumn{3}{c}{\textbf{MATH}}            \\
                                     \cmidrule(l{3mm}r{3mm}){2-4} \cmidrule(l{3mm}r{3mm}){5-7}
\textbf{Models}                      & \textbf{Accuracy$\uparrow$} & \textbf{Tokens$\downarrow$} & \textbf{CCA$\uparrow$} & \textbf{Accuracy$\uparrow$} & \textbf{Tokens$\downarrow$} & \textbf{CCA$\uparrow$} \\
\midrule
Qwen2.5-Math       &  93.21     &  439.04     & 14.8 &  62.27    & 577.94       & 10.1 \\
Qwen2.5-Math-RS    &  97.17     &  193.73     & 23.5 &  73.01    & 395.31       & 17.6 \\
Qwen2.5-Math-\ours    &  \textbf{97.59}  & \textbf{97.01} & \textbf{36.0} & \textbf{73.87} & \textbf{288.24} & \textbf{24.3} \\
\midrule
DS-Distill-Qwen    & 94.72   & 455.61       & 21.0 & 70.33  & 1006.94          & 13.2 \\
DS-Distill-Qwen-RS & 97.06   & 497.88       & 33.7 & \textbf{82.87}  & 557.12  & 25.8 \\
DS-Distill-Qwen-\ours & \textbf{97.72}  & \textbf{100.58} & \textbf{52.2} & 82.77 & \textbf{270.51} & \textbf{38.6}\\
\bottomrule
\end{tabular}%

\caption{
Performance of models on GSM8k and MATH datasets. Each model is trained and evaluated on the same dataset, and then assessed by three metrics. 
``RS'' indicates that the base model is fine-tuned by raw score.
``\ours'' denotes that \ours~reward is applied to the base model during the training process.
``Qwen2.5-Math'' and ``DS-Distill-Qwen'' are the abbreviations of ``Qwen2.5-Math-7B'' and ``DeepSeek-R1-Distill-Qwen-7B''.
}
\label{tab:main_results}

\end{table*}

\begin{table*}[htbp]
\resizebox{\textwidth}{!}{%
\begin{tabular}{lcccccccc}
\toprule
                   & \multicolumn{2}{c}{\textbf{AIME24}} & \multicolumn{2}{c}{\textbf{AMC24}} & \multicolumn{2}{c}{\textbf{CNMO24}} & \multicolumn{2}{c}{\textbf{GPQA}} \\
                    \cmidrule(l{3mm}r{3mm}){2-3} \cmidrule(l{3mm}r{3mm}){4-5} \cmidrule(l{3mm}r{3mm}){6-7} \cmidrule(l{3mm}r{3mm}){8-9}
\textbf{Models}    & \textbf{Accuracy$\uparrow$} & \textbf{Tokens$\downarrow$} & \textbf{Accuracy$\uparrow$} & \textbf{Tokens$\downarrow$} & \textbf{Accuracy$\uparrow$} & \textbf{Tokens$\downarrow$} & \textbf{Accuracy$\uparrow$} & \textbf{Tokens$\downarrow$} \\
\midrule
Qwen2.5-Math     & 6.7  & 1868.50  &  4.3  & 1509.89   &  11.1 & 1072.56  & 30.1 & 847.97  \\
Qwen2.5-Math-RS  & \textbf{26.7}  & 751.63  &  21.7  & 729.52   &  \textbf{22.2} & 847.22  & \textbf{32.3} & 639.25  \\
Qwen2.5-Math-\ours  & \textbf{26.7}  & \textbf{421.51}  &  \textbf{23.9}  & \textbf{395.10}   &  \textbf{22.2} & \textbf{546.67}  & 30.0 & \textbf{568.01}  \\
\midrule
DS-Distill-Qwen    & 16.7 & 3218.06 &  26.1 & 3142.20 &  16.7 & 3108.72 & 13.1 & 3054.42 \\
DS-Distill-Qwen-RS & 26.7  & 3038.67 & 47.8  & 1598.24 &  61.1 & 2887.39 & 32.3 & 3002.72 \\
DS-Distill-Qwen-\ours & \textbf{30.0}  & \textbf{2724.07} &  \textbf{52.2}  & \textbf{1264.65} & \textbf{66.6} & \textbf{2729.83} & \textbf{37.9} & \textbf{2749.09} \\
\bottomrule
\end{tabular}%
}
\caption{
Evaluate models on out-of-distribution math datasets.
}
\label{tab:evaluation_results}
\vspace{-5mm}
\end{table*}

\subsection{Dynamic Schedule}
\label{sec:scheduling}

The parameter $\beta$ in the length reward determines when the penalty starts to take effect, but it relies on the target accuracy $A_\text{target}$. A well-calibrated $A_\text{target}$ is essential to reduce overthinking while preserving model performance.
A naive approach is to manually set $A_\text{target}$ and apply the length reward only after the model surpasses a fixed accuracy or training step threshold. However, this method is impractical, as neither the optimal accuracy nor the convergence point is known beforehand.
Inspired by the momentum~\citep{QIAN1999145} method used in learning rate scheduling algorithms, we propose two dynamic length reward scheduling strategies that adaptively set $A_\text{target}$ during training.




\paragraph{Exponential Moving Average (EMA).}
This method updates the target accuracy by smoothing it toward the best validation accuracy seen so far:
\begin{align*}
    A_{\text{val}}^{\text{max}} &= \max_{0 < i \leq t} A_{\text{val}, i} \\
    A_{\text{target}, i} &= \max\big( \epsilon \cdot A_{\text{target}, i-1} + (1 - \epsilon) \cdot A_{\text{val}}^{\text{max}}, \nonumber \\
    &\quad\quad\quad\;\;\; A_{\text{val}}^{\text{max}} \big)
\end{align*}
where $t$ is the current step index, and the $\epsilon$ controls the inertia of the target. The initial value of $A_{\text{target}, i}$ is 1.
When validation accuracy exceeds the previous target, the target is directly updated to prevent lag.

\paragraph{Potential Scheduling (PS).}
This strategy models the target accuracy as the current best validation accuracy plus a decaying potential:
\begin{align*}
    A_{\text{val}}^{\text{max}} &= \max_{0 < i \leq t} A_{\text{val}, i} \\
    P_i &= 
    \begin{cases}
    1 - A_{\text{val}, 0} & i = 0 \\
    \min\left(1 - A_{\text{val}}^{\text{max}},\ \epsilon \cdot P_{i-1} \right) & i > 0
    \end{cases} \\
    A_{\text{target}, i} &= A_{\text{val}}^{\text{max}} + P_i
\end{align*}
where $t$ is the same as the above method, and $\epsilon$ controls how fast the potential decays, smaller values reduce $P_i$ more aggressively over time.
The $\min$ function ensures $A_{\text{target},i} \leq 1$ even when accuracy improves rapidly.
This method ensures that $A_\text{target}$ always stays above the best validation accuracy while gradually reducing the potential gap.

\section{Experiments}

\subsection{Settings}

\paragraph{Models and Datasets}
We conduct our training experiments on two commonly used and well-known models: Qwen2.5-Math-7B~\cite{yang2024qwen25mathtechnicalreportmathematical} and DeepSeek-R1-Distill-Qwen-7B~\cite{deepseekai2025deepseekr1incentivizingreasoningcapability}, which have shown excellent reasoning ability on various math tasks.
Both models are trained on the GSM8k~\citep{cobbe2021training} and MATH~\cite{hendrycksmath2021} datasets. Both training set has about 7.5k questions.
During our training, we split both training sets in 95\%:5\% ratio to create validation sets for the training process.
We evaluate our models on different reasoning datasets: GSM8k, MATH, AIME24, AMC24, CNMO24, and GPQA~\cite{rein2024gpqa}.
We fine-tune models using the GRPO methods. The default parameters during the training process are as follows.
The values of $\alpha$ and $\beta$ are $10^{-6}$ and $128$, respectively.
Both values of $\gamma$ and $\epsilon$ are 0.9.
We choose the first dynamic scheduling method as the de fault training method.
The length limitation of both prompts and generations is 1k tokens, which is long enough for most of our questions and answers in training datasets.
Models are fine-tuned for about 300 steps to reach convergence on both inference length and accuracy.
All experiments are conducted on the AWS EC2 platform using 8 NVIDIA H100 GPUs.

\paragraph{Baselines}
To evaluate the superiority of our method, we compare our results against two baselines.
(1) Original models: are published models without any further fine-tuning.
(2) Original models-RS: are fine-tuned original models by raw scores (0/1 scores). If an answer extracted from the inference text is correct, the raw score is 1. Otherwise, it is 0.

\paragraph{Metrics}
We evaluate using accuracy, average inference length, and Consistent Concise Accuracy (CCA)~\cite{nayab2024concise}, a composite metric evaluate both efficiency and accuracy. For CCA, we follow their settings of $\alpha=10$ and $\beta=40$, and using $k=1024$ in our experiments. All metrics are averaged over 3 runs.



\subsection{Results}

Table \ref{tab:main_results} demonstrates the performance of various methods on different models. Models trained with our rewards answer questions in short lengths, while slightly improving accuracy.

\begin{figure}[htbp]
    \centering
    \includegraphics[width=0.95\columnwidth]{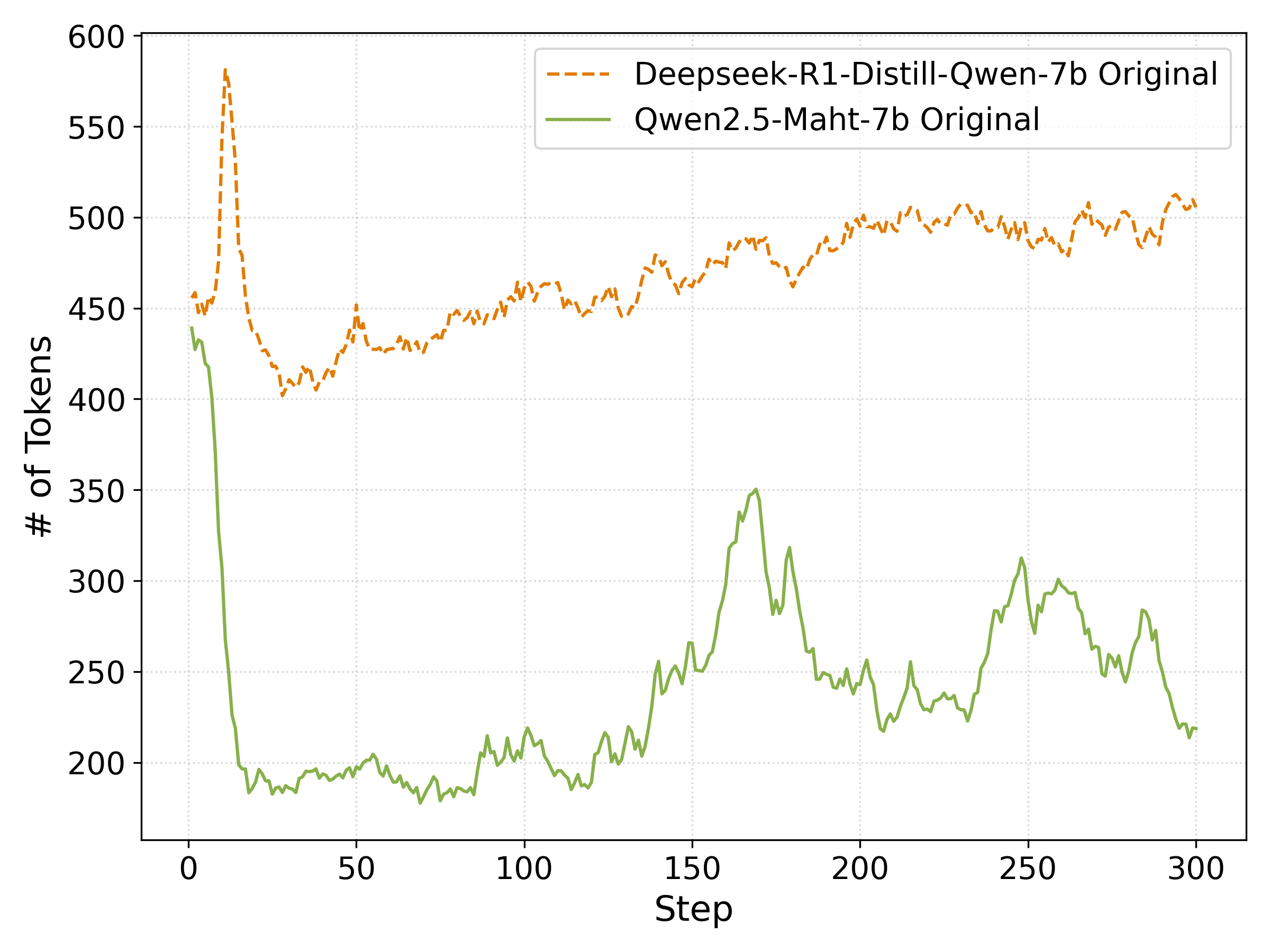}
    \vspace{-5mm}
    \caption{\# of inference tokens of models during training with raw scores.
    }
    \label{fig:tokens-compare-two-model-ori}
    \vspace{-3mm}
\end{figure}

\paragraph{Performance Analysis}
Notably, results demonstrate that our length reward framework can dramatically prevent overthinking while improving models' reasoning ability compared to the original model.
On the GSM8k and MATH datasets, models finetuned with \ours~only need less than a quarter and a half of the original length to generate correct answers, respectively, and show a significant improvement in CCA.
Moreover, applying the raw score (0/1 score) during training also causes the model outputs to gradually decrease in most cases. This is mainly because post-training can slightly simplify outputs by removing redundant content. Detailed analysis of the change of inference texts will be discussed in section \ref{sec:analysis}.
Furthermore, the training process on the GSM8k dataset is unstable as shown in Figure \ref{fig:tokens-compare-two-model-ori}. It is mainly because the dataset is simple, and post-training by raw score as the reward cannot effectively simplify it.

\paragraph{Evaluation on Out-of-Distribution Datasets}
In addition to the above two datasets, we also evaluate our models trained on the MATH dataset, on out-of-distribution datasets, as shown in Table \ref{tab:evaluation_results}.
The original models perform poorly due to constraints on generation length. Most outputs are truncated by the length limitation rule and do not contain correct answers.
In contrast, models trained with \ours~consistently outperform both the original models and those trained with raw scores across all datasets, while also generating fewer tokens.
Thus, it demonstrates that our length reward improves models' generalizability and robustness on out-of-distribution complex datasets.

\paragraph{Scaling Analysis}

\begin{figure}[htbp]
    \centering
    \begin{minipage}{\columnwidth}
        \centering
        \includegraphics[width=0.9\columnwidth]{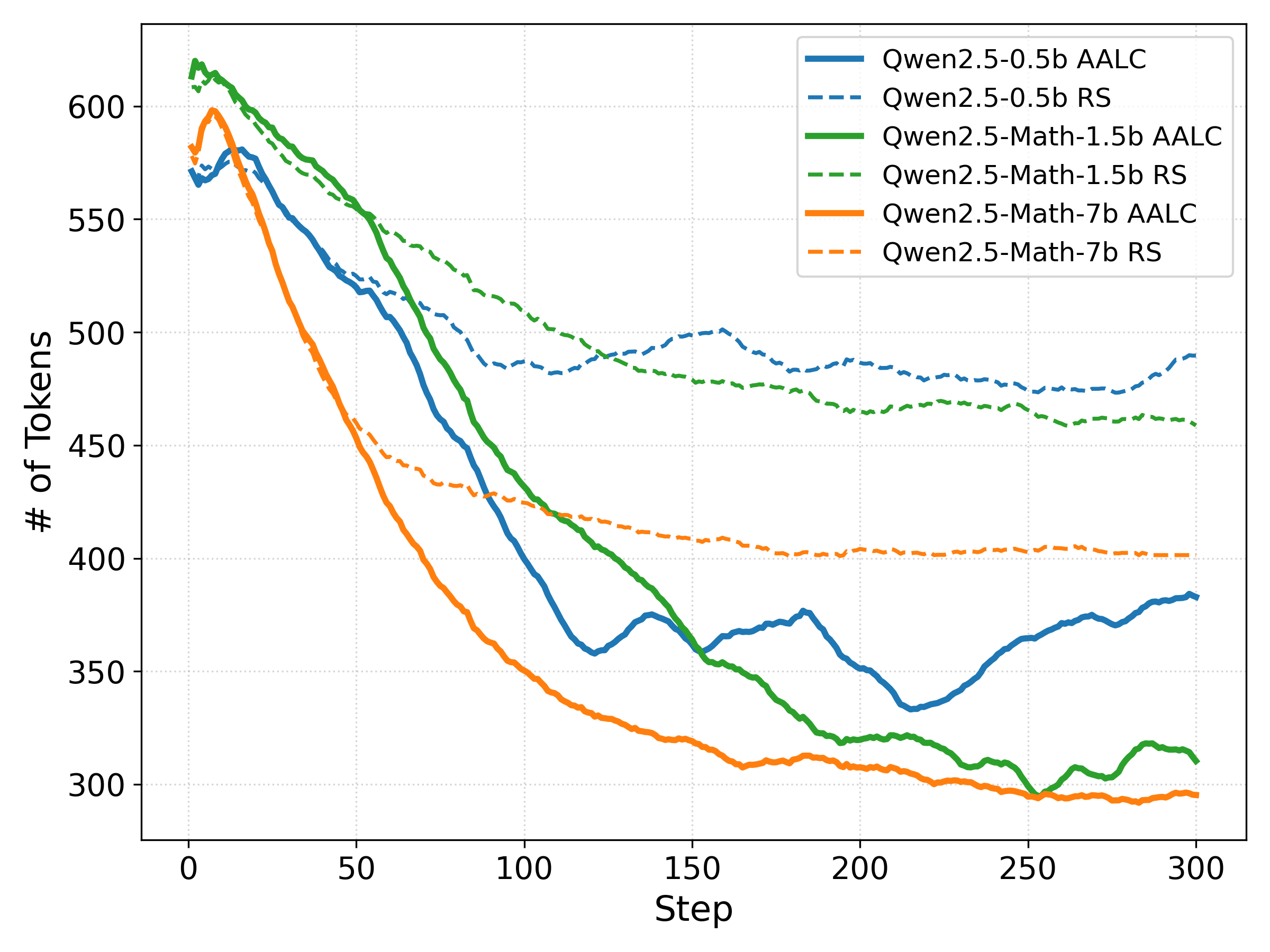}
        \vspace{-4mm}
        \caption{
            Training Progress of Qwen2.5 Models Across Scales. Bold lines indicate training runs that incorporate the \ours. Models of the same scale are represented with similar colors.
        }
        \label{fig:scaling}
    \end{minipage}

    \vspace{3mm}

    \begin{minipage}{\columnwidth}
        \centering
        \begin{tabular}{l|cc}
            \toprule
            \textbf{Models} & \textbf{Acc (RS)} & \textbf{Acc (\ours)} \\
             \midrule
            Qwen2.5-0.5b        & \textbf{34.0} & 33.9 \\
            Qwen2.5-Math-1.5b   & 67.8 & \textbf{68.4} \\
            Qwen2.5-Math-7b     & 73.0 & \textbf{73.9} \\
             \bottomrule
        \end{tabular}
        \vspace{-2mm}
        \captionof{table}{
            Accuracy comparison of Qwen2.5 models trained by Raw Score (RS) vs. \ours~across scales. 
        }
        \label{tab:scaling}
    \end{minipage}
    \vspace{-5mm}
\end{figure}

As shown in Figure~\ref{fig:scaling} and Table~\ref{tab:scaling}, our \ours~reward method consistently improves training across various model scales. All training runs incorporating \ours~effectively mitigate overthinking while preserving accuracy compared to those using only raw scores (RS).
Since, the smallest official model in the Qwen2.5-Math family is 1.5B, we include Qwen2.5-0.5B as a substitute to examine behavior at a smaller scale, 
Although its training is less stable due to limited reasoning capability, it still follows the same performance trend observed in larger models, validating the generality of our approach.




\subsection{Ablation Study}

\label{sec:ablation-study}

\begin{figure*}[ht]
    \vspace{-2mm}
    \centering
    \begin{subfigure}[b]{0.3\textwidth}
        \includegraphics[width=\textwidth]{\detokenize{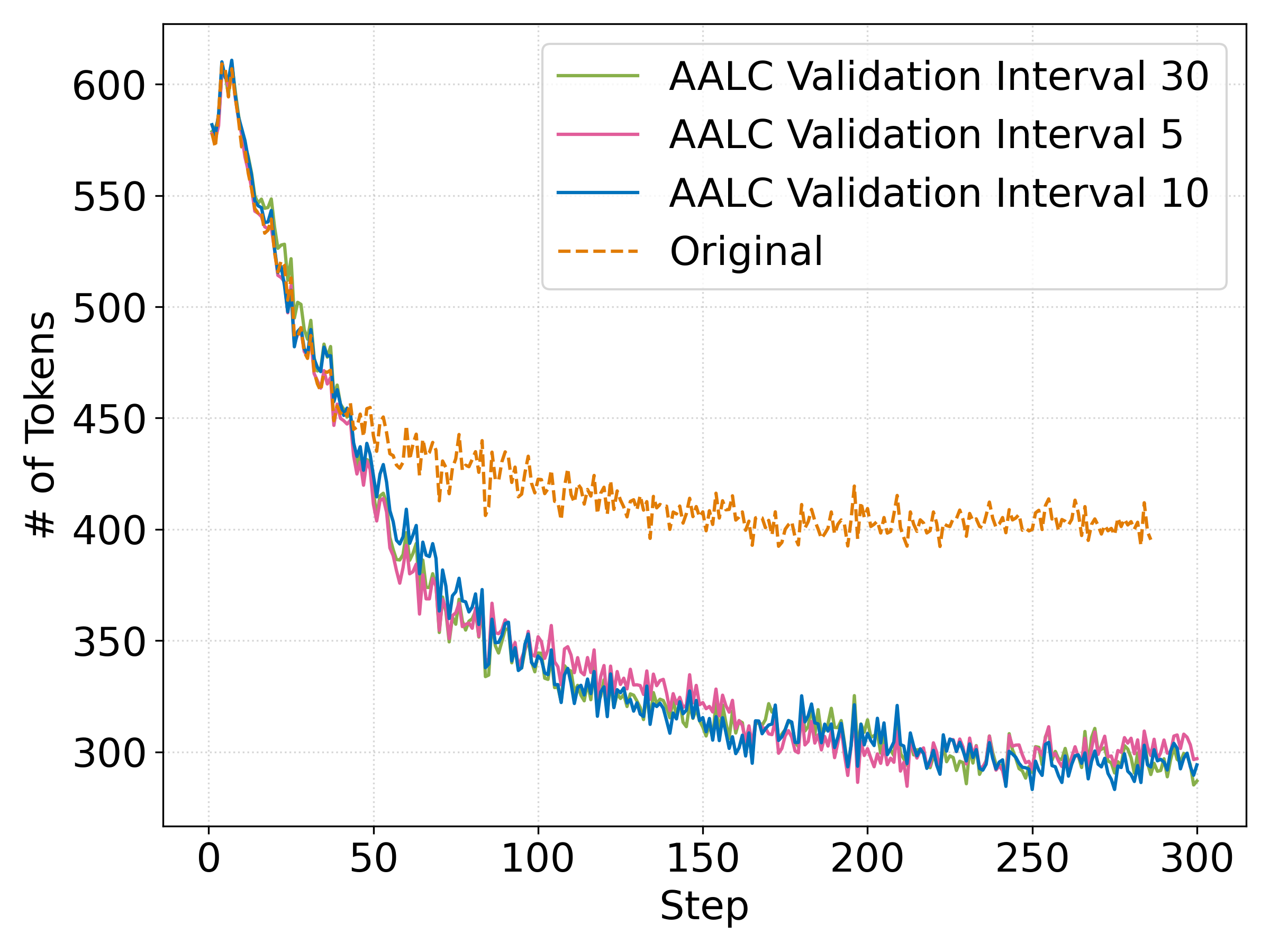}}
        \caption{The output length at different validation steps.}
        \label{fig:ablation-length-val-interval}
    \end{subfigure}
    \hfill
    \begin{subfigure}[b]{0.3\textwidth}
        \includegraphics[width=\textwidth]{\detokenize{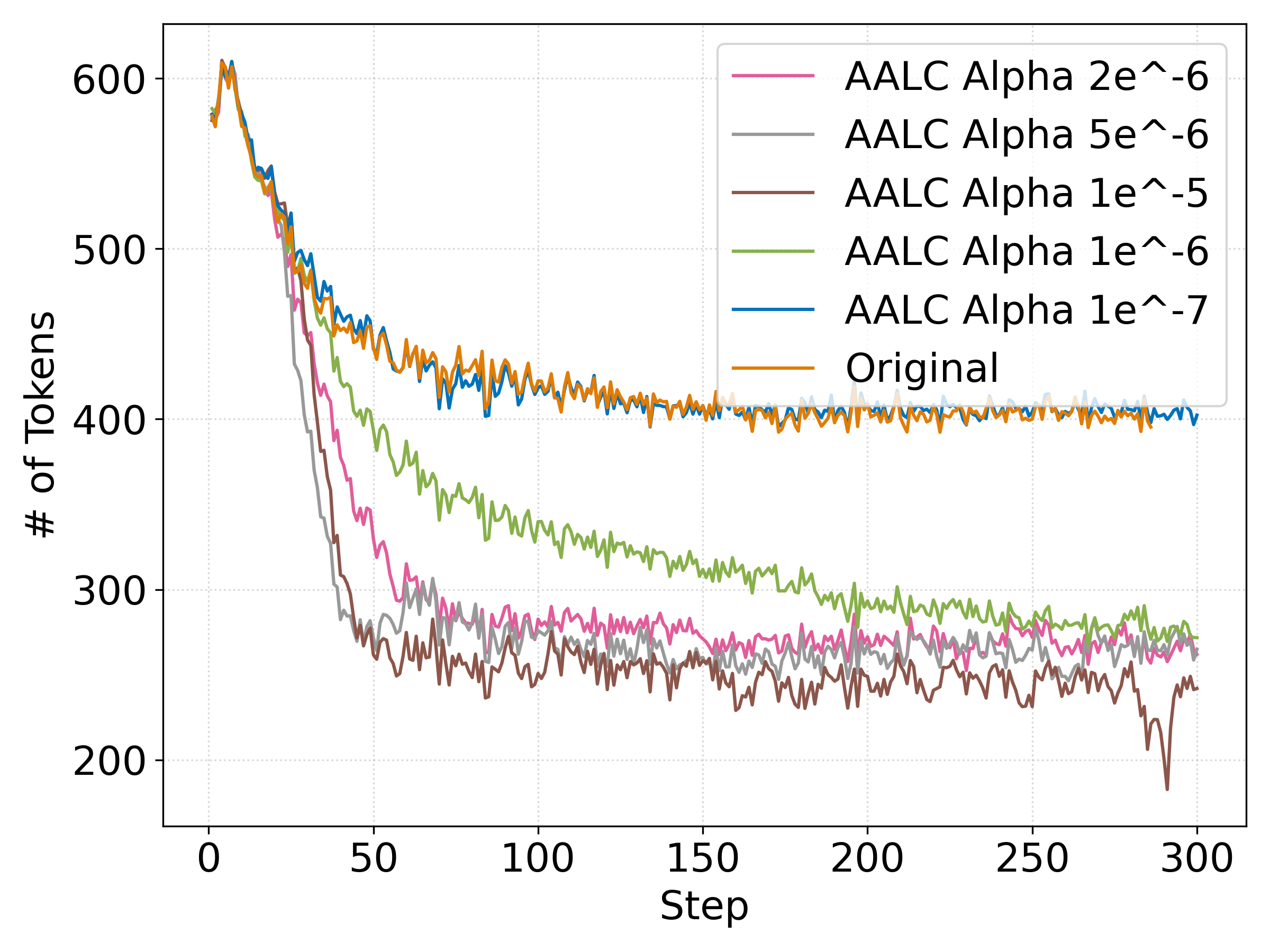}}
        \caption{The output length of different $\alpha$ values.}
        \label{fig:ablation-length-alpha}
    \end{subfigure}
    \hfill
    \begin{subfigure}[b]{0.3\textwidth}
        \includegraphics[width=\textwidth]{\detokenize{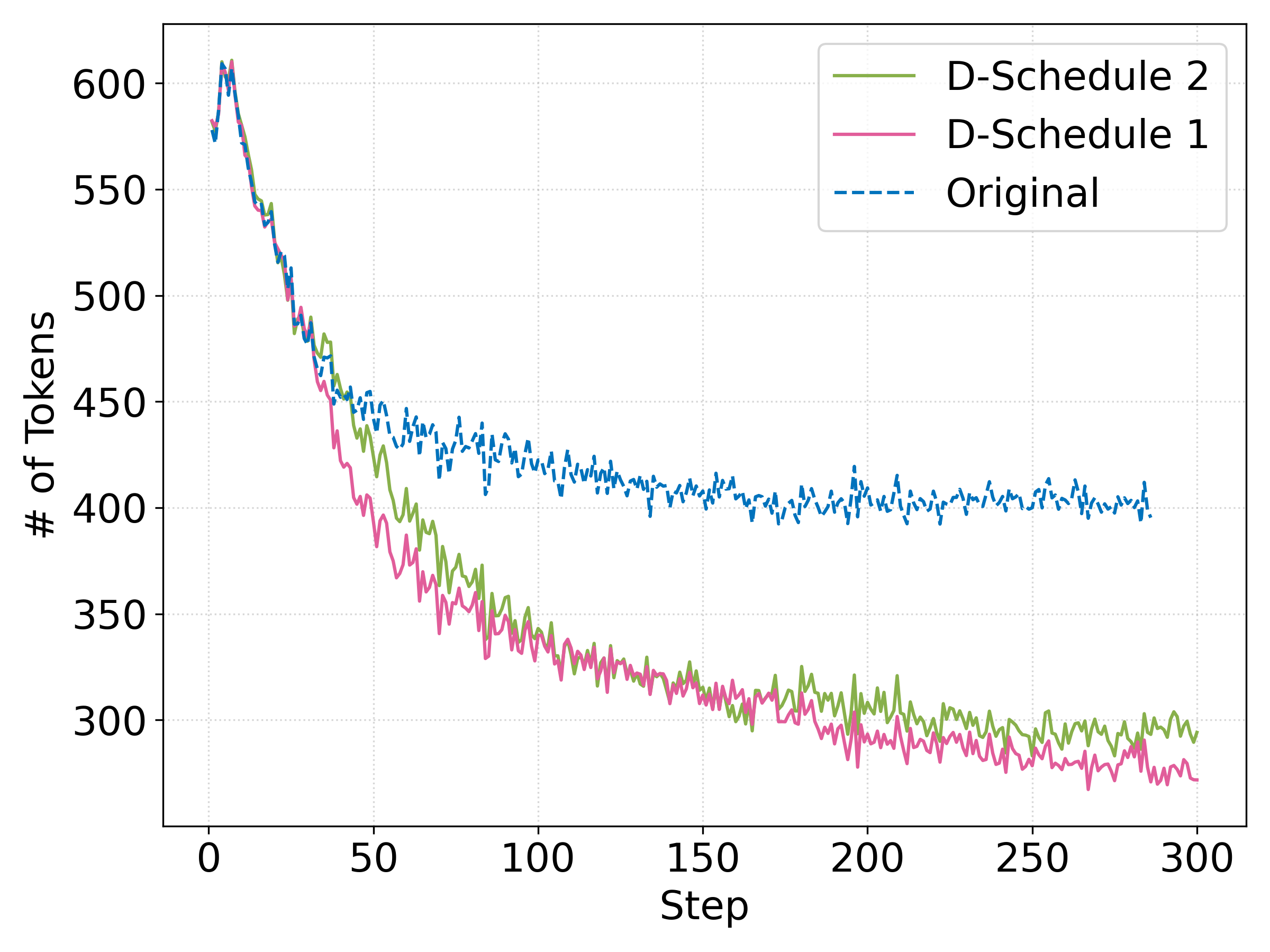}}
        \caption{The output length of different dynamic schedule methods.}
        \label{fig:ablation-length-dschedule}
    \end{subfigure}

    \vspace{0.2cm}

    \begin{subfigure}[b]{0.3\textwidth}
        \includegraphics[width=\textwidth]{\detokenize{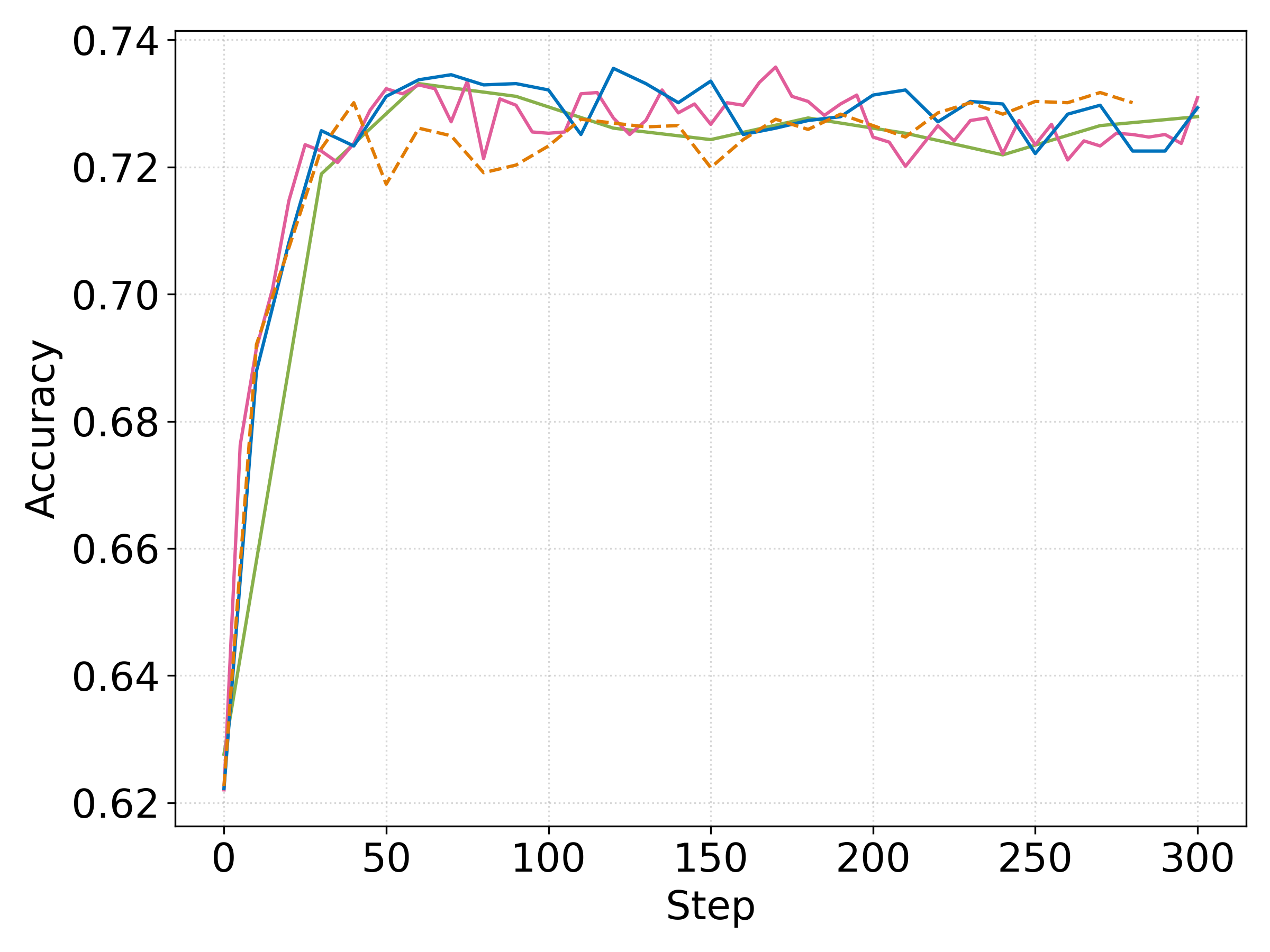}}
        \caption{The output accuracy at different validation steps.}
        \label{fig:ablation-accuracy-val-interval}
    \end{subfigure}
    \hfill
    \begin{subfigure}[b]{0.3\textwidth}
        \includegraphics[width=\textwidth]{\detokenize{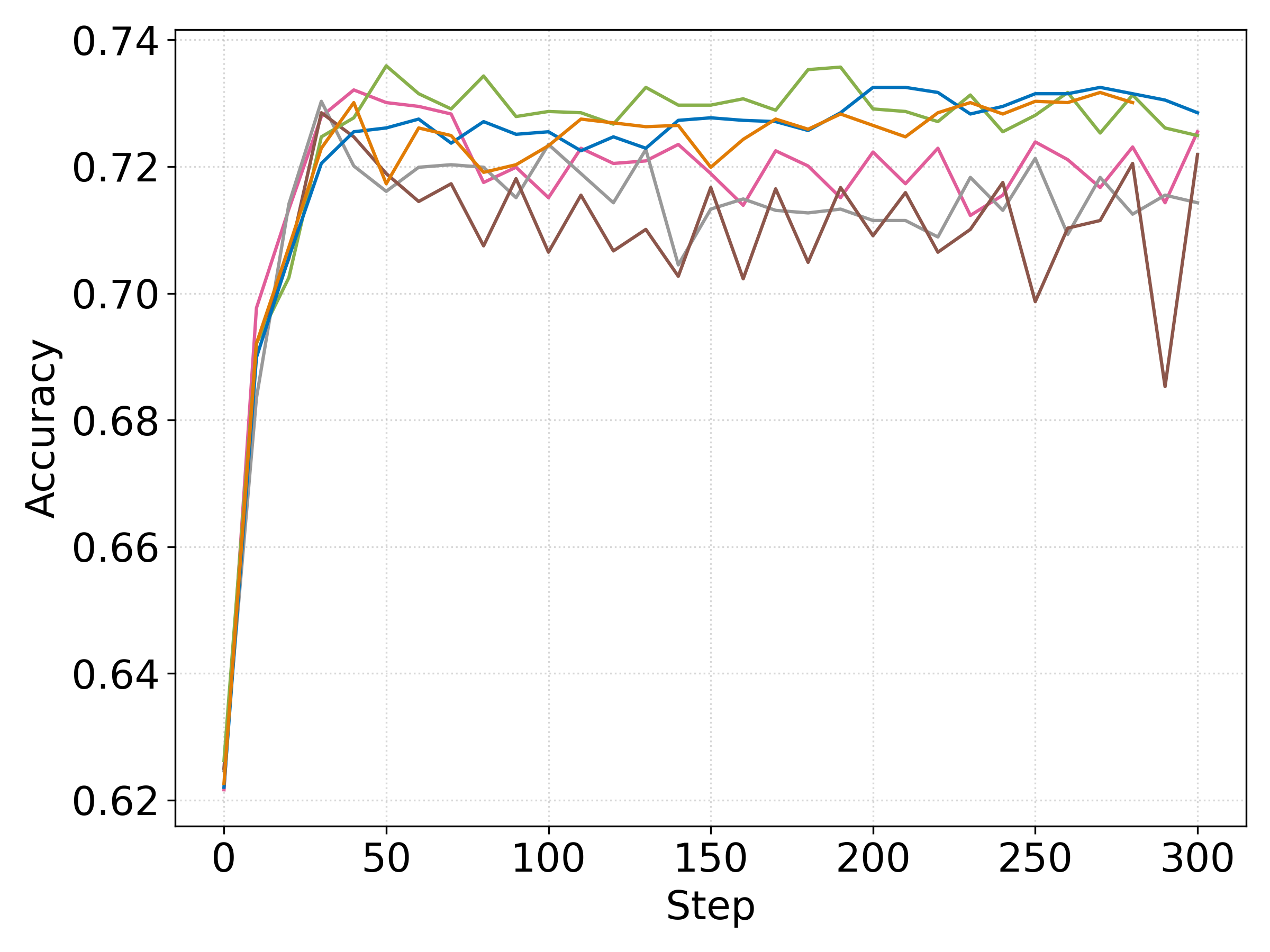}}
        \caption{The output accuracy of different $\alpha$ values.}
        \label{fig:ablation-accuracy-alpha}
    \end{subfigure}
    \hfill
    \begin{subfigure}[b]{0.3\textwidth}
        \includegraphics[width=\textwidth]{\detokenize{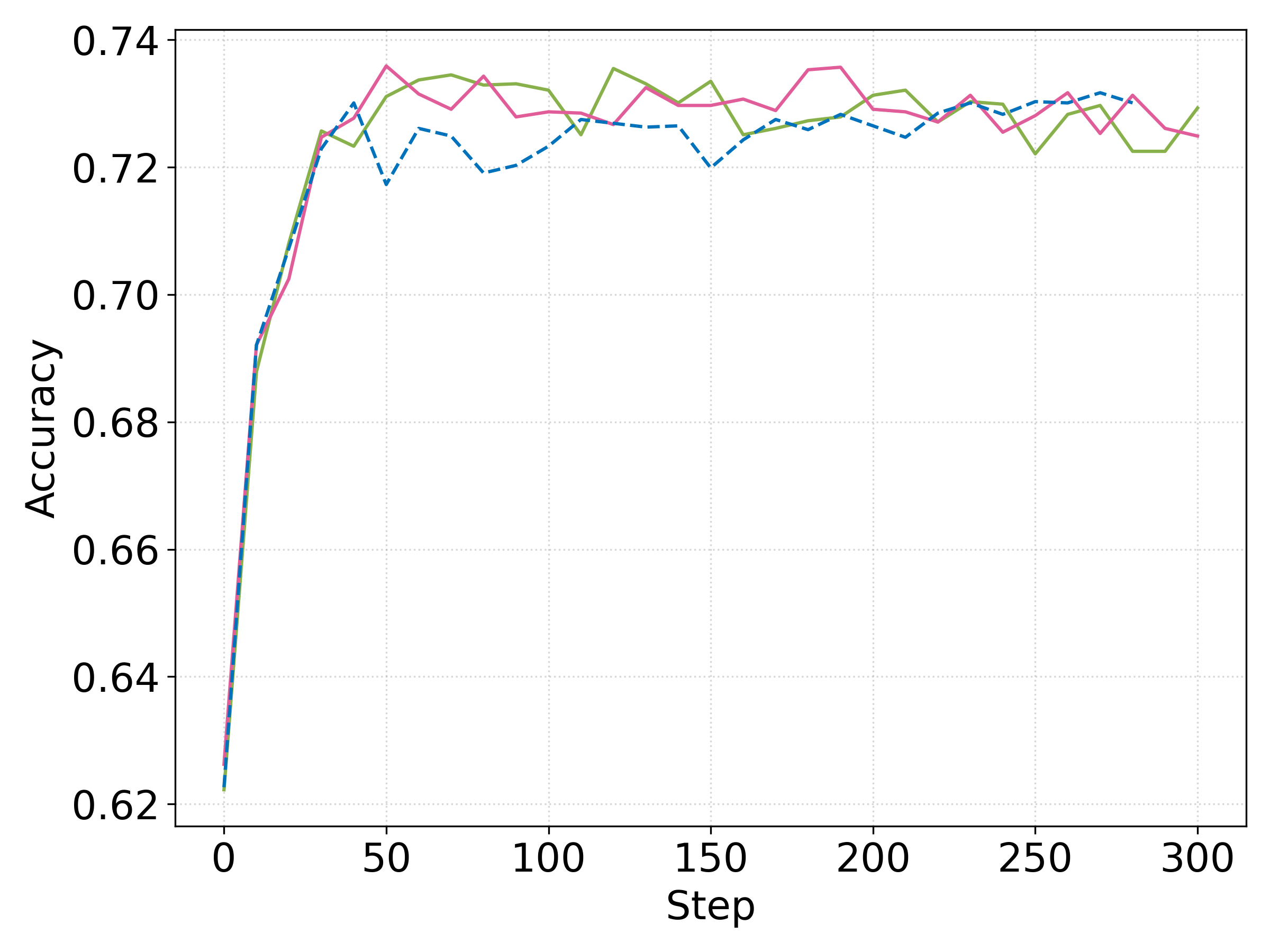}}
        \caption{The output accuracy of different dynamic schedule methods.}
        \label{fig:ablation-accuracy-dschedule}
    \end{subfigure}

    \vspace{-2mm}
    \caption{
    Ablation analysis of token usage and accuracy under different training configurations. 
    Top row (a–c): Impact on output length (token count) when varying validation step interval (a), length reward weight $\alpha$ (b), and different dynamic schedule strategy (c). 
    Bottom row (d–f): Corresponding effects on validation accuracy for the same settings. 
    }
    \label{fig:token_usage_steps}
\end{figure*}

\paragraph{Hyper Parameters $\alpha$, $\beta$ $\gamma$, and $\epsilon$}
We run all the following experiments following the second dynamic schedule method on Qwen2.5-Math-7B model training on the MATH dataset.
The baseline settings are the same as the ``Length Penalty'' setting in the main result section, which is $\alpha=1e^{-6}$, $\beta=128$, $\gamma=0.9$, and $\epsilon=0.9$.

We evaluate five values of $\alpha$: $10^{-5}$, $5\times 10^{-6}$, $2\times 10^{-6}$, $10^{-6}$, and $10^{-7}$ (see Figures~\ref{fig:ablation-length-alpha} and~\ref{fig:ablation-accuracy-alpha}). Among them, $\alpha = 10^{-6}$ yields the best trade-off, effectively reducing overthinking while preserving high accuracy. A smaller value like $10^{-7}$ fails to curb verbosity, while a larger value such as $10^{-5}$ leads to a notable drop in accuracy (from 73\% to 71\%) with only a modest reduction of $\sim 30$ tokens in output length. Notably, the training curve for $\alpha = 10^{-5}$ shows greater fluctuation than others; both accuracy and length decline sharply toward the end of training before recovering. This pattern highlights the auto-adaptability of our dynamic schedule: when accuracy degrades, the method prioritizes restoring performance over enforcing brevity.

\paragraph{Dynamic Schedule}
We compare our dynamic scheduling methods under the same hyperparameter settings as shown in the above section. Figure \ref{fig:ablation-length-dschedule} and \ref{fig:ablation-accuracy-dschedule} show that the two methods are similar in effectiveness. However, the first dynamic scheduling method converges faster and more stably at the end of training.

\paragraph{Validation Interval}
Since validation accuracy is an essential part of formulas and controls two parts of the framework, the frequency of obtaining the value may be vital to the training process. We have tried three interval values: $5$, $10$, and $30$. Results are in Figure \ref{fig:ablation-length-val-interval} and \ref{fig:ablation-accuracy-val-interval}
Experiments show that there is almost no difference in choosing different validation intervals.
With a small validation interval, the reduction of generation length is smoother than with others. However, it does not affect the final results.
More detailed ablation study results are in Appendix C.

\section{Analysis}
\label{sec:analysis}

\subsection{Semantic Quality Analysis}
\begin{table*}[ht]
\centering
\begin{tabular}{lllllll}
\toprule
                               &                                & \multicolumn{2}{c}{GSM8k} &  & \multicolumn{2}{c}{MATH} \\ \cline{3-4} \cline{6-7} 
\multicolumn{1}{l}{Our methods}      & Opponent                       & Human       & GPT-4o      &  &   Human          &   GPT-4o         \\
\midrule
Qwen2.5-Math-\ours             & Qwen2.5-Math                &     64.4\%        &          55.9\%   &   &            66.8\%  &    57.2\%        \\
                               & Qwen2.5-Math-RS          &    60.5\%  &   53.4\%           &          & 58.3\%  &  53.6\%                         \\ \hline
DS-Distill-Qwen-\ours & DS-Distill-Qwen    &            61.3\% &    52.8\%         &  &  67.5\%           &    54.6\%        \\
                               & DS-Distill-Qwen-RS &          57.8\%   &   52.5\%          &  &    61.6\%           &    54.1\%            \\
\bottomrule
\end{tabular}
\caption{Win rates of our proposed method against various baselines settings on GSM8k and MATH benchmark. Evaluations include human annotators and GPT-4o.
}
\label{tab:winrate}
\vspace{-5mm}
\end{table*}

In this section, in addition to standard metrics such as accuracy and response length, we evaluate the semantic quality of reduced responses, providing a complementary dimension for assessing efficiency in reasoning tasks.
Specifically, we present the win rate of our proposed method compared to other baseline in Table \ref{tab:main_results}. This evaluation includes both both manual assessments and automated comparisons using GPT-4o~\citep{hurst2024gpt}. During these pairwise comparisons, each model response earns 1 point for a win, 0.5 points for a tie, and 0 points for a loss. The win rate is thus calculated as the proportion of total points earned by our method relative to the baseline method across all comparisons. Given a problem instance $x$ with corresponding solutions $y_1$ and $y_2$, the evaluation criteria are as follows:
\begin{itemize}[itemsep=0pt,topsep=0pt,parsep=0pt,leftmargin=*]
\item If $y_1$ is correct and $y_2$ is not, $y_1$ wins; if $y_2$ is correct and $y_1$ is not, $y_2$ wins; 
\item If both $y_1$ and $y_2$ are incorrect, neither receives points;
\item If both are correct, their semantic quality is evaluated by human or GPT-4o to determine the superior one.
\end{itemize}
The evaluation prompt template, detailed in Appendix B, 
is specifically designed to mitigate length bias during semantic quality assessment. To avoid positional bias, the order of $y_1$ and $y_2$ is randomized. This evaluation approach is motivated by the fact that, in reasoning tasks, only correct solutions are semantically meaningful; incorrect solutions lack value regardless of their conciseness or perceived faithfulness.

The results of this analysis are presented in Table \ref{tab:winrate}. Our approach consistently achieves a win rate above 50\% against all baselines.
Moreover, we observe that human evaluations yield a higher win rate compared to GPT-4o assessments, likely due to GPT-4o's known preference bias towards lengthier responses~\citep{singhal2023long}. Despite a slight reduction in raw accuracy scores compared to baseline methods shown in Table \ref{tab:main_results}, our method demonstrates superior semantic quality. 
These findings validate our method’s ability to generate concise yet semantically richer solutions.

\subsection{Reasoning Behavior Analysis}
\label{sec:behavior analysis}

To understand the changes of model’s reasoning patterns throughout the training process, following~\citep{zeng2025simplerl}, we adopt the cognitive behavior framework proposed by~\citet{gandhi2025cognitive}, leveraging GPT-4o~\citep{hurst2024gpt} to identify four behaviors: “Backtracking”, “Verification”, “Subgoal Setting”, and “Enumeration”. This approach is more robust compared to traditional keyword-based methods~\citep{xie2025logic, yeo2025demystifying}, effectively capturing nuanced reasoning patterns even without explicit textual indicators. Definitions and evaluation prompts for these behaviors are provided in Appendix A.

\begin{figure}[h]
    \centering
    \includegraphics[width=0.9\linewidth]{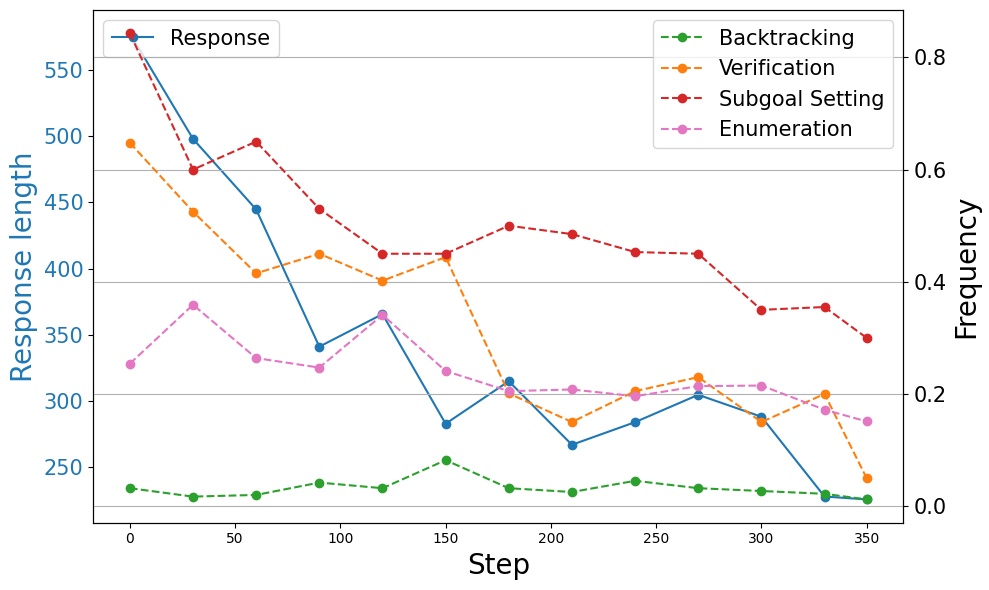}
    \vspace{-2mm}
    \caption{Response length and frequency of reasoning behaviors identified in Qwen2.5-Math-\ours~on MATH.}
    \label{fig:qwen-reflection}
    \vspace{-2mm}
\end{figure}

\begin{figure}[htbp]
    \centering
    \includegraphics[width=0.9\linewidth]{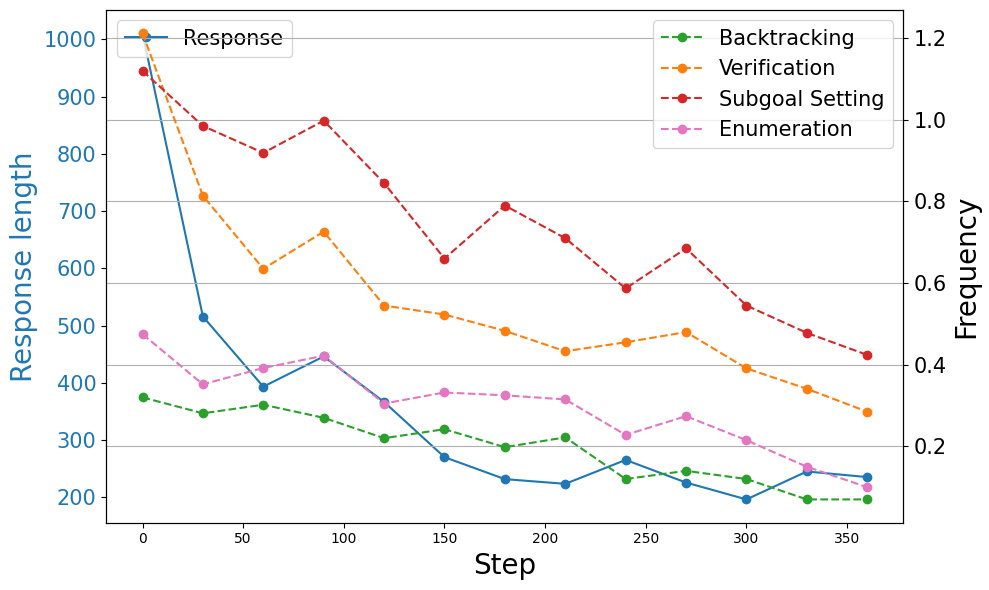}
    \vspace{-2mm}
    \caption{Response length and frequency of reasoning behaviors identified in DS-Distill-Qwen-\ours~on MATH.}
    \label{fig:deepseek-reflection}
    \vspace{-4mm}
\end{figure}

Specifically, following the experiment setting in Table \ref{tab:main_results}, we conduct this reasoning behavior analysis on two models, Qwen2.5-Math-\ours~and DS-Distill-Qwen-\ours~on the MATH benchmark. 
Figures \ref{fig:qwen-reflection} and \ref{fig:deepseek-reflection} show the correlation between response length and the frequency of identified reasoning behaviors during training. Our analysis reveals that ``Subgoal Setting” and ``Verification” are predominant, significantly contributing to reasoning verbosity. As training progresses, we observe a concurrent reduction in the frequency of all identified reasoning behaviors, aligning closely with decreased output lengths. These observations suggest that our method for improving reasoning efficiency is primarily achieved by reducing redundant reasoning behavior.
We also present more qualitative analysis in Appendix D.

\subsection{More Qualitative Analysis}
\label{sec:5.3}

To understand the impact of our \ours reward on inference behavior, we compare the Qwen2.5-Math models trained with Raw Score with a variant trained with \ours(with a learning rate of 1e-6). A case study is provided in Table~\ref{tab:final-vs-dlp} of Appendix \ref{app:case study}. While both models yield correct answers, we observe that the length-penalized model significantly compresses explanations and eliminates scaffolding, presenting solutions in a more direct and streamlined manner. This behavior is consistent across both the MATH and GSM8k datasets. 
Structural elements such as framing phrases, code-style reasoning, and boxed outputs are removed entirely. 
These shorter, higher-confidence completions often result in improved exact match accuracy. 
We attribute these gains to the model’s preference for more direct solution paths under length constraints, which suppresses narrative redundancy and promotes structurally compressed, low-variance reasoning.

\paragraph{More Insights}
Further analysis of early- vs. mid-training stages and base vs. distilled models  reveals three key insights (Appendix~\ref{app:More Qualitative Analysis}):
First, the effects of length penalty emerge very early (by training step 60, see Appendix \ref{app:start working}). 
Second, gains in efficiency coincide with a loss of interpretability: model outputs increasingly omit narrative framing, explanatory context, and derivation steps (see Table \ref{tab:step30-vs-step60}).
Third, when \ours~is applied to both a base model and its distilled version (e.g., Qwen2.5-Math and DeepSeek-R1-Distill-Qwen), the distilled model’s outputs converge to lengths comparable to the base model’s while preserving accuracy (see Appendix~\ref{app:distill}). 
These findings suggest that the benefit of reduced output length  arise from structural refinements in reasoning process rather than mere truncation.






\section{Conclusion}

In this work, we present a lightweight, accuracy-aware length reward mechanism to mitigate overthinking in large language models, promoting concise yet accurate reasoning. By dynamically balancing brevity and correctness within a reinforcement learning framework, our method consistently reduces response length, often by over 50\%, while preserving or even improving performance. These results hold consistently across model families, scales, and out‐of‐distribution datasets, highlighting the generality of our approach. Analyses reveal that our approach suppresses redundant cognitive behaviors, such as excessive subgoal setting and verification, and induces structural refinement rather than mere truncation. These findings demonstrate the potential of reward-based strategies to guide LLMs toward more efficient and robust reasoning paths.

\bibliography{custom}
\bibliographystyle{acl_natbib}

\newpage
\appendix
\onecolumn


\section{Reasoning Behavior Analysis}
\label{app:Reasoning Behavior Analysis}
\citet{gandhi2024stream} identify four core cognitive behaviors displayed by language models:

\begin{enumerate}
\item[(1)] \textbf{Verification}: systematically checking intermediate results for correctness;
\item[(2)] \textbf{Backtracking}: detecting errors mid-generation and explicitly revising prior steps;
\item[(3)] \textbf{Subgoal setting}: decomposing a complex problem into smaller, tractable sub-tasks;
\item[(4)] \textbf{Enumeration}: exhaustively considering multiple cases or possibilities when searching for a solution.
\end{enumerate}

We substitute “Enumeration” for the original “Backward Chaining,” which is less pertinent to mathematical-reasoning tasks.

\begin{tcolorbox}[colback=white,colframe=black,title=Prompt Template for Identifying and Analyzing Reasoning Behaviors]

Below is a chain-of-reasoning generated by a Language Model when attempting to solve a math problem. Evaluate this chain-of-reasoning to determine whether it demonstrates beneficial problem-solving behaviors that deviate from typical linear, monotonic reasoning patterns commonly observed in language models.

\medskip

\texttt{<start\_of\_reasoning>} \\
\texttt{\{input\}} \\
\texttt{<end\_of\_reasoning>}

\medskip

Specifically, actively identify and emphasize beneficial behaviors such as:
\begin{itemize}
    \item \textbf{Backtracking}: Explicitly revising approaches upon identifying errors or dead ends (e.g., \textit{"This approach won't work because..."}).
    \item \textbf{Verification}: Systematically checking intermediate results or reasoning steps (e.g., \textit{"Let's verify this result by..."}).
    \item \textbf{Subgoal Setting}: Breaking down complex problems into smaller, manageable steps (e.g., \textit{"To solve this, we first need to..."}).
    \item \textbf{Enumeration}: Solving problems by exhaustively considering multiple cases or possibilities.
\end{itemize}

Additionally, remain attentive to and encourage the identification of other beneficial behaviors not explicitly listed here, such as creative analogies, abstraction to simpler cases, or insightful generalizations.

\medskip

\textbf{Important:} \\
Clearly specify each beneficial behavior you identify. \\
Provide explicit examples from the reasoning chain. \\
If no beneficial behaviors are observed, explicitly return an empty list. \\
Provide your evaluation clearly, formatted as follows:

\medskip
\definecolor{codebg}{RGB}{255, 255, 240}

\begin{tcolorbox}[colback=codebg, colframe=gray!50, boxrule=0.5pt]
\ttfamily
\begin{tabbing}
\hspace{2cm} \= \kill
\{ \\
\> "behavior": "", \\
\> "example": "" \\
\}
\end{tabbing}
\end{tcolorbox}
\end{tcolorbox}

\newpage

\subsection{Qwen3-235B-A22B Output}
\begin{tcolorbox}[title=Qwen Output, colback=white, colframe=blue!60!black]
We are given:

- Gloria can either buy \textbf{a pair of boots} or \textbf{two pairs of high heels}.
- The \textbf{total cost of the two pairs of heels is \$5 less than the cost of the boots}.
- One pair of heels costs \$33, and the other costs twice as much.

\textbf{Step 1:} Total cost of heels:
\[
33 + (2 \times 33) = 33 + 66 = 99
\]

\textbf{Step 2:} Cost of boots:
\[
99 + 5 = \boxed{104}
\]
\end{tcolorbox}

\subsection{ChatGPT-4o Output}
\begin{tcolorbox}[title=Qwen Output, colback=white, colframe=blue!60!black]
We are given:
\begin{itemize}
    \item Gloria can either buy a \textbf{pair of boots} or \textbf{two pairs of high heels}.
    \item The \textbf{total cost of the two pairs of heels is \$5 less than the cost of the boots}.
    \item One pair of heels costs \$33, and the other costs \textbf{twice as much}.
\end{itemize}

\textbf{Step 1: Define variables} \\
Let:
\begin{itemize}
    \item \( B \) = cost of the \textbf{boots}
    \item The first pair of heels = \$33
    \item The second pair of heels = \$2 x 33 = \$66
\end{itemize}

So, total cost of both pairs of heels =
\[
33 + 66 = 99
\]

We're told this is \$5 less than the boots, so:
\[
B = 99 + 5 = 104
\]

\textbf{\color{green!50!black} Final Answer:} \\
The boots cost \textbf{\$104}.
\end{tcolorbox}

\subsection*{ChatGPT-4o Output}
\begin{tcolorbox}[title=ChatGPT Output, colback=white, colframe=green!50!black]

\noindent Let's solve the problem \textbf{step by step}.

\vspace{1em}
\hrule
\vspace{1em}

\textbf{Step 1: Understand the problem}

Gloria finds a pair of \textbf{boots} that cost a certain amount. \\
She also considers buying \textbf{two pairs of high heels} instead.

\begin{itemize}
    \item The \textbf{first pair} of heels costs \$33.

    \item The \textbf{second pair} of heels costs \textbf{twice as much}:
    \[
    2 \times 33 = 66\ \text{dollars}
    \]

    \item The \textbf{total cost of the two heels} is:
    \[
    33 + 66 = 99\ \text{dollars}
    \]

    \item The \textbf{boots cost \$5 more} than the two pairs of heels:
    \[
    \text{Boots} = 99 + 5 = 104\ \text{dollars}
    \]
\end{itemize}

\vspace{1em}
Final Answer:
\[
\boxed{104}
\]

\end{tcolorbox}

\subsection*{Gemini-2.0 Output}
\begin{tcolorbox}[title=Gemini Output, colback=white, colframe=red!60!black]
Let's break this down step by step:

\begin{enumerate}[label=\textbf{\arabic*.}]
    \item \textbf{Cost of the more expensive heels:} One pair of heels costs \$33, and the other costs twice as much. So, the more expensive heels cost \$33 \(\times\) 2 = 66.
    
    \item \textbf{Total cost of the two pairs of heels:} To find the combined cost of both pairs of heels, we add their individual prices: \$33 + 66 = 99.
    
    \item \textbf{Cost of the boots:} The problem states that the two pairs of high heels together cost five dollars less than the boots. This means the boots cost \$5 more than the combined cost of the heels. So, the boots cost \$99 + 5 = 104.
\end{enumerate}

\vspace{0.5em}
Therefore, the boots cost \fbox{\textbf{104}} dollars.
\end{tcolorbox}

\subsection*{DeepSeek Output}
\begin{tcolorbox}[title=DeepSeek Output, colback=white, colframe=orange!70!black]
Let's determine the cost of the boots step by step.
\begin{enumerate}[leftmargin=*, label=\textbf{\arabic*.}]
    \item \textbf{Identify the Costs of the High Heels:}
    \begin{itemize}
        \item \textbf{First pair of heels:} \$33
        \item \textbf{Second pair of heels:} Twice as much as the first pair.
        \[
        2 \times \$33 = \$66
        \]
    \end{itemize}

    \item \textbf{Calculate the Total Cost of Both Pairs of Heels:}
    \[
    \$33\ (\text{first pair}) + \$66\ (\text{second pair}) = \$99
    \]

    \item \textbf{Determine the Cost of the Boots:}
    \begin{itemize}
        \item The problem states that the two pairs of heels together cost \textbf{\$5 less} than the boots.
        \item Let \( B \) be the cost of the boots.
    \end{itemize}
    \[
    \$99 = B - \$5
    \]
    \[
    B = \$99 + \$5 = \$104
    \]
\end{enumerate}

\noindent \textbf{Final Answer:}

\[
\boxed{104}
\]
\end{tcolorbox}

\newpage

\newpage

\section{Win Rate Analysis}
\label{app:Rate Analysis}
We use the following prompt template for win rate analysis with three criteria that helps GPT-4o keep a focus on instruction following and semantic quality while avoiding length bias:

\begin{tcolorbox}[colback=white, colframe=black, title=\textbf{Response Semantic Quality Comparison Template}]
Given the following instruction and two candidate responses:

Instruction: {\color{red}\{\}} \\
Response 1: {\color{red}\{\}} \\
Response 2: {\color{red}\{\}} \\

Please evaluate which response better follows the instruction based on the following criteria: \\
\hspace*{1em}1. Adherence to the instruction. \\
\hspace*{1em}2. Overall semantic quality. \\
\hspace*{1em}3. Conciseness: the response should \textbf{avoid} redundancy, unnecessary verbosity, or inclusion of irrelevant information.

Based on your evaluation, respond in the format: \\
\textit{'Response 1 is better than Response 2', 'Response 2 is better than Response 1', or 'Response 1 is equal to Response 2'}.
\end{tcolorbox}

\section{More Ablation Study}
\label{app:more-ablation-study}

\begin{figure*}[htbp]
    \vspace{-5mm}
    \centering
    \begin{minipage}[b]{0.49\textwidth}
        \centering
        \includegraphics[width=\textwidth]{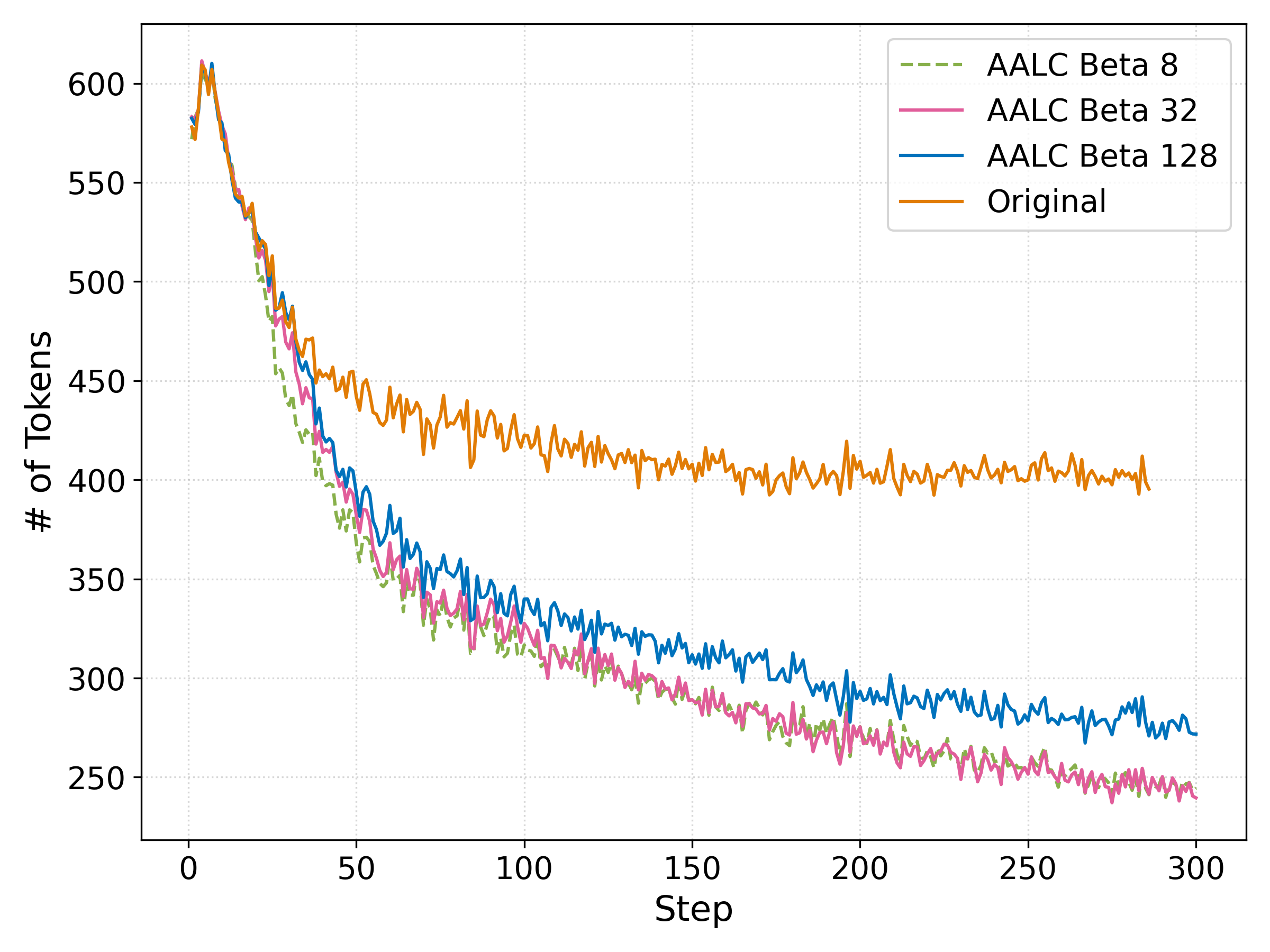}
        \caption{The output length at different $\beta$ values.}
        \label{fig:ablation-length-beta}
    \end{minipage}\hfill
    \begin{minipage}[b]{0.49\textwidth}
        \centering
        \includegraphics[width=\textwidth]{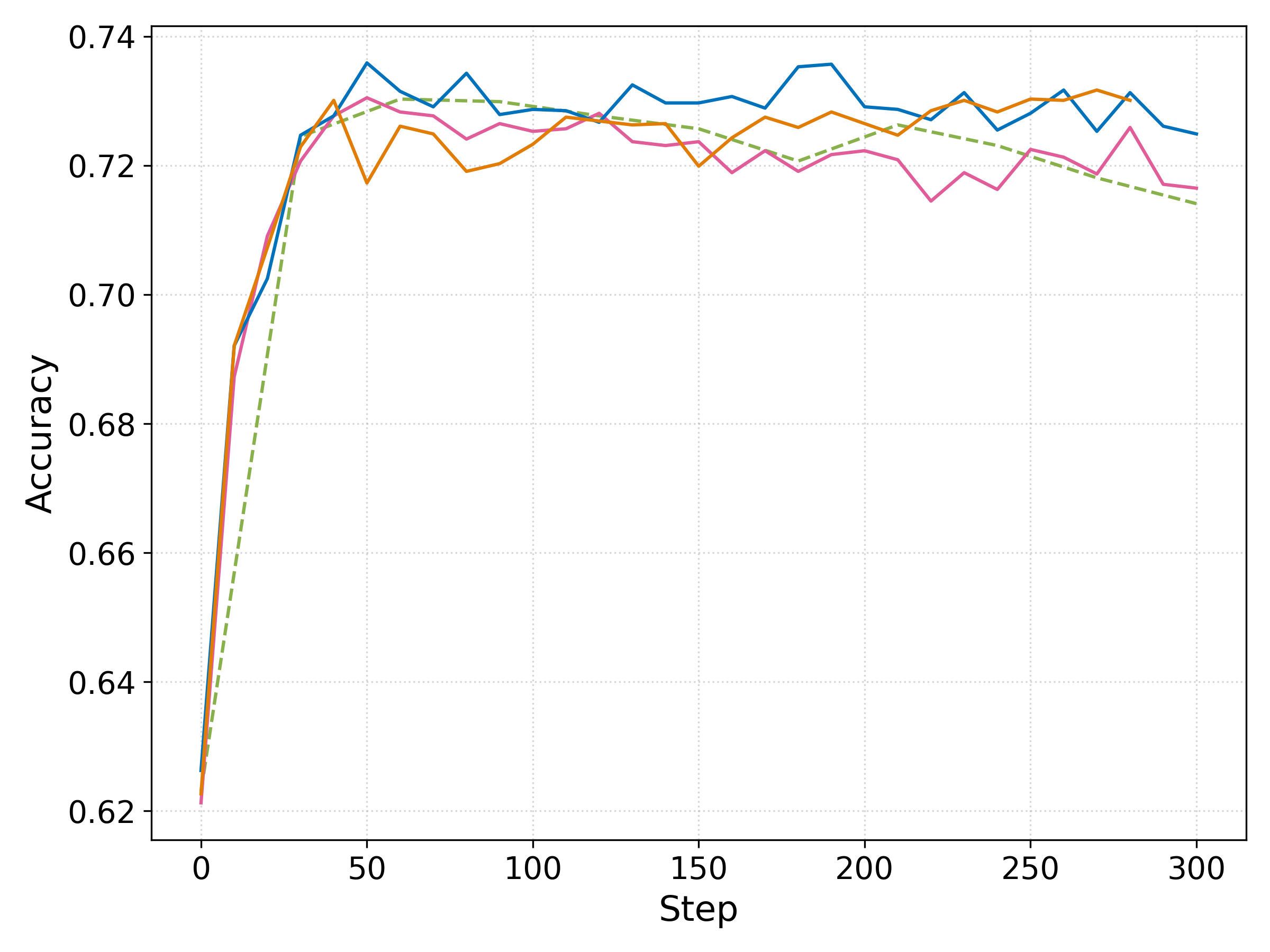}
        \caption{The output accuracy at different $\beta$ values.}
        \label{fig:ablation-accuracy-beta}
    \end{minipage}
\end{figure*}

For $\beta$, we have tried the following values: 8, 32, and 128. Results are in the figures \ref{fig:ablation-length-beta} and \ref{fig:ablation-accuracy-beta}. The $\beta$ controls when the length reward intervenes in the training process.
With a low $\beta$ value, such as 8, the length reward starts being effective early as the validation accuracy reaches around 45\% of the target accuracy. According to Figure \ref{fig:ablation-length-beta}, early engagement in the training process can reduce the generation length more, but the model's performance also decreases from 73\% to 71\%.
With a high $\beta$ value, such as 128, the length reward does not affect the training process until the validation accuracy reaches 95\% of the target accuracy. The validation accuracy process also shows the model's performance is even better than the ``Original'' training process.
In our framework, our goal is to maintain the model's performance while preventing overthinking. Thus, a high value is preferred in this case. 


\newpage

\section{More Qualitative Analysis}
\label{app:More Qualitative Analysis}


\subsection{Case Study: Raw Score vs. \ours}
\label{app:case study}
To demonstrate the effect of \ours, as shown in Table~\ref{tab:final-vs-dlp}, given one problem each from the MATH and GSM8K datasets, we compare responses from the Qwen2.5-Math model trained with raw scores versus those trained with \ours.
The results indicate that the former produces detailed problem interpretations and symbolic derivations, whereas the latter reduces verbose commentary while preserving essential computations.

\begin{table*}[htbp]
    \centering
    \renewcommand{\arraystretch}{1.3}
\begin{adjustbox}{width=\textwidth}
\begin{tabular}{|p{4cm}|p{6cm}|p{6cm}|}
\hline
\textbf{Dataset / Question} & \textbf{Output of Raw Score} & \textbf{Output of \ours} \\
\hline

\textbf{MATH:} \newline \textit{How many vertical asymptotes does the graph of $y=\frac{2}{x^2+x-6}$ have?}
& \textit{To determine the number of vertical asymptotes of the function \( y = \frac{2}{x^2 + x - 6} \), we need to find the values of \( x \) where the denominator is zero.}\newline
\textit{The denominator is \( x^2 + x - 6 = (x + 3)(x - 2) \). So, the solutions are \( x = -3 \) and \( x = 2 \).}\newline
\textit{These are the vertical asymptotes. Final answer: \( \boxed{2} \).}
& \textit{The function $y=\frac{2}{x^2+x-6}$ has vertical asymptotes where the denominator is zero. The denominator $x^2 + x - 6$ factors as $(x+3)(x-2)$. So, the asymptotes are at $x = -3$ and $x = 2$. Final answer: $\boxed{2}$.} \\
\hline

\textbf{GSM8k:} \newline \textit{A robe takes 2 bolts of blue fiber and half that much white fiber. How many bolts in total does it take?}
& \textit{The robe takes 2 bolts of blue fiber and half that, i.e., 1 bolt, of white fiber. Total = $2 + 1 = 3$.} \newline
& \textit{2 blue bolts + 1 white bolt = $3$ bolts in total. Final answer: $\boxed{3}$.} \\
\hline
\end{tabular}
\end{adjustbox}
\caption{Compare the outputs of Qwen2.5-Math model trained with raw score and that trained with dynamic length penalty ($\alpha = 1e{-6}$).}
\label{tab:final-vs-dlp}
\end{table*}

\begin{table*}[t]
\centering
\renewcommand{\arraystretch}{1.3}
\begin{tabular}{|p{3.8cm}|p{6.1cm}|p{6.1cm}|}
\hline
\textbf{Dataset / Question} & \textbf{Step 30 Checkpoint Output} & \textbf{Step 60 Checkpoint Output} \\
\hline

\textbf{Math:} \newline \textit{What is the sum of the first 10 positive even numbers?}
& Detailed explanation: defines sequence, uses formula \( S = \frac{n}{2}(a + l) \), explains each term, then computes: \newline
\texttt{S = 10/2 * (2 + 20) = 5 * 22 = \boxed{110}} \newline
Includes full derivation and justification (not shown here).
& Condensed version: \newline
\textit{``First 10 even numbers: 2 to 20. Sum = 10/2 × (2 + 20) = 5 × 22 = 110. \#\#\#\# 110''} \newline
No sequence explanation or setup. \\
\hline

\textbf{GSM8K:} \newline \textit{Kylar wants to buy 16 glasses. One costs \$5; every second glass costs 60\%.}
& Full arithmetic + Python code: \newline
\texttt{cost\_first = 5, cost\_second = 5 * 0.6} \newline
\texttt{total\_cost = (cost\_first + cost\_second) * 8} \newline
\texttt{print(total\_cost) \# Output: 64.0} \newline
Explains pairing logic and includes validation (omitted here).
& Shortened to only arithmetic steps: \newline
\textit{``8 cheaper glasses cost 8 * \$3 = \$24. 8 full-price glasses cost 8 * \$5 = \$40. Total = 64. \#\#\#\# 64''} \newline
All narrative and code removed. \\
\hline
\end{tabular}
\caption{Comparison between early (Step 30) and mid-stage (Step 60) model outputs. Length penalty begins to reduce verbosity while preserving core reasoning accuracy.}
\label{tab:step30-vs-step60}
\end{table*}

\subsection{When Does the \ours~Start Working? A Step 30 vs Step 60 Analysis}
\label{app:start working}
To investigate when the \ours~begins to meaningfully affect model behavior, we compare inference outputs at step 30 and step 60 under \ours~settings (1e-6 learning rate, dynamic schedule) on both the {MATH} and {GSM8k} datasets. These checkpoints are chosen to align with the transition period of the d-schedule mechanism, which gradually increases the weight of the length penalty during post-training. By step 60, we already observe clear signs of compression: outputs are more concise, pedagogical scaffolding is reduced, and verbose explanation structures are stripped away. Table~\ref{tab:step30-vs-step60} illustrates this behavioral shift with representative examples.

At step 30, the model still retains a “teacher-like” reasoning style. For example, in response to a math problem asking for the sum of the first 10 positive even numbers, the model describes the arithmetic sequence, defines variables explicitly, applies the formula \( S = \frac{n}{2}(a + l) \), and concludes with a boxed answer. In contrast, the step 60 output removes the setup entirely and provides the final arithmetic chain directly: \textit{``Sum = 10/2 × (2 + 20) = 110. \#\#\#\# 110''}. A similar transition is observed in the GSM8k example: at step 30, the model walks through glass pricing logic with Python-like code and a \texttt{print()} statement for validation; by step 60, this is reduced to three compact sentences with no code, commentary, or explanation.

These early changes follow a consistent pattern. Narrative framing—phrases like “Let’s solve this by...” or “Now compute...” — are dropped first. Embedded symbolic or code-based verification disappears shortly after. While the arithmetic logic is preserved, the accompanying verbal scaffolding is eliminated. This compression begins after the d-schedule’s early burn-in phase (typically around step 20–40), as the model starts receiving stronger training signals to minimize token count while retaining correctness. By step 60, the pressure from the length penalty is sufficiently strong to shape model behavior measurably.

In summary, the model begins transitioning to a compressed reasoning style as early as step 60. The resulting outputs are more efficient and better aligned with inference-time brevity goals, but less transparent in their reasoning. This supports the hypothesis that the d-schedule enforces compression gradually, with tangible effects emerging soon after step 30. While answer correctness remains intact, the interpretability of the solution path is the first to be sacrificed.

\subsection{How Distillation Improves Reasoning Quality}
\label{app:distill}
\begin{figure}[h]
    \centering
    \begin{subfigure}[t]{0.48\linewidth}
        \centering
        \includegraphics[width=\linewidth]{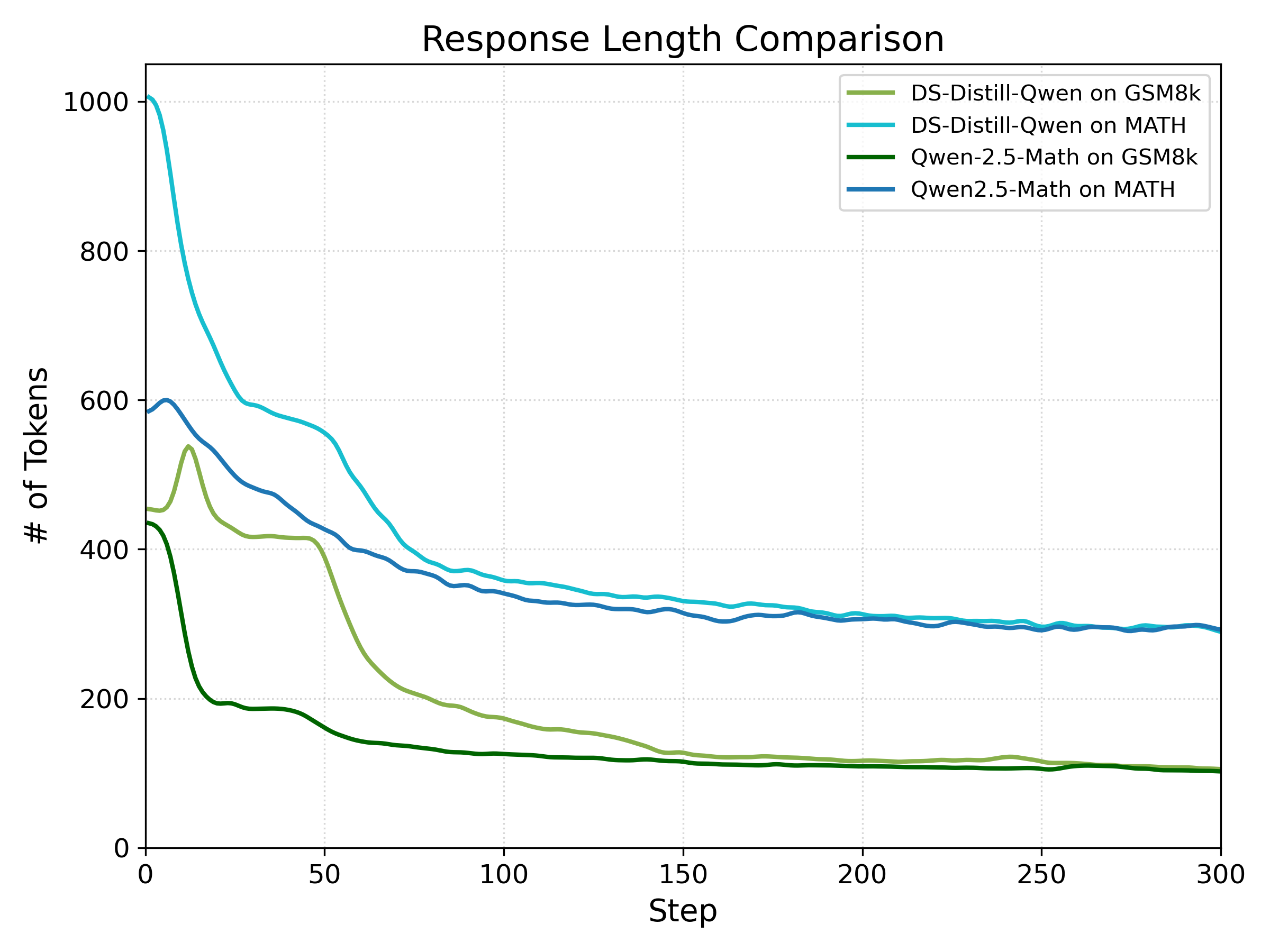}
        \caption{Response length over steps.}
        \label{fig:sub-response-length}
    \end{subfigure}
    \hfill
    \begin{subfigure}[t]{0.48\linewidth}
        \centering
        \includegraphics[width=\linewidth]{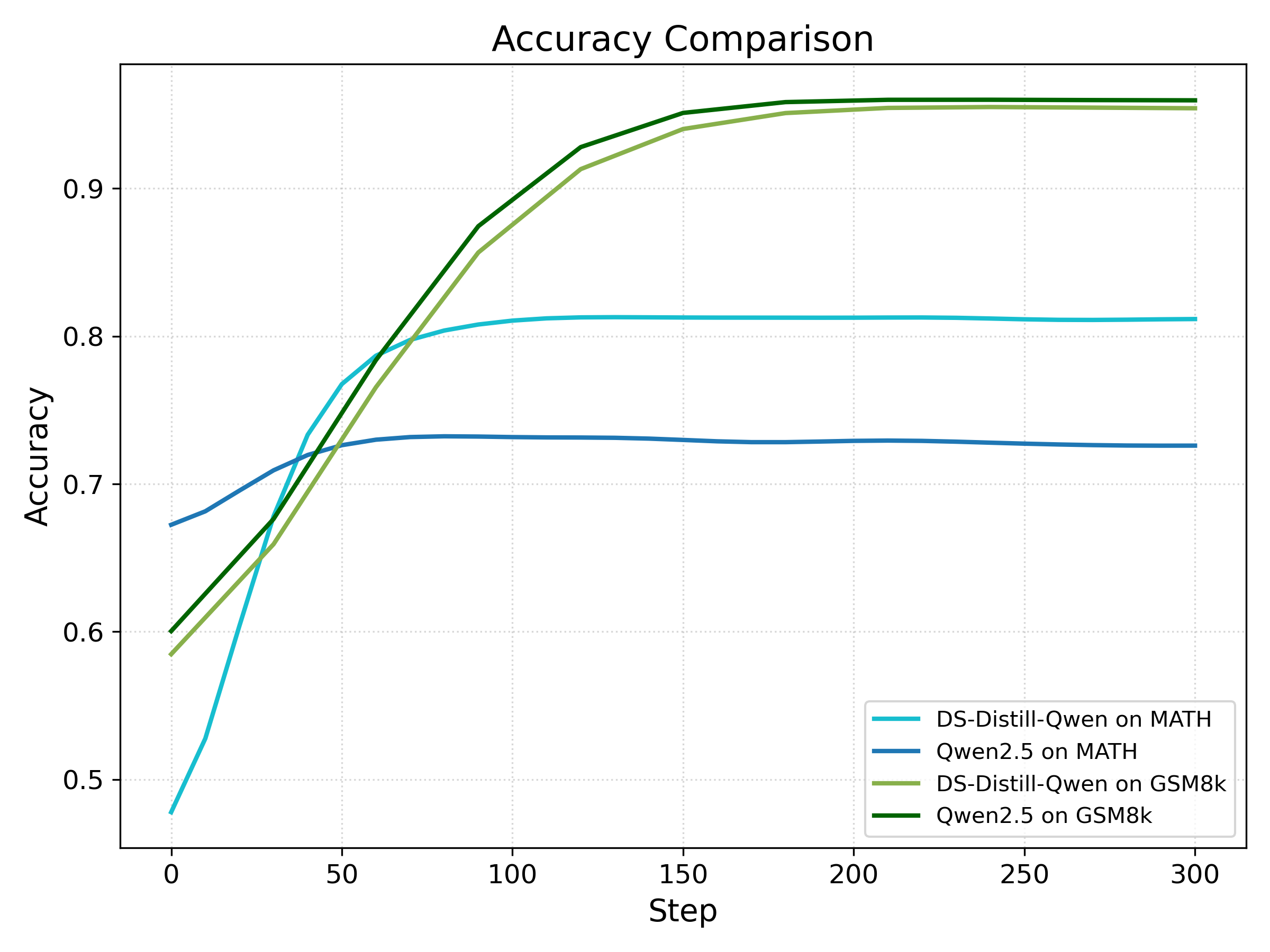}
        \caption{Accuracy over steps.}
        \label{fig:sub-accuracy}
    \end{subfigure}
    \caption{(a) Comparison of response lengths over training steps between DeepSeek-Distill-Qwen and Qwen2.5 models on GSM8k and MATH benchmarks. DeepSeek-Distill-Qwen consistently generates longer outputs, especially in earlier training phases. (b) Accuracy comparison over training steps between DeepSeek-Distill-Qwen and Qwen2.5 on GSM8k and MATH datasets. DeepSeek-Distill-Qwen achieves higher accuracy on both benchmarks. All model outputs are capped at 1000 tokens for fair comparison.}
    \label{fig:combined-comparison}
    \vspace{-3mm}
\end{figure}

We applied \ours~on both Deepseek-R1-Distill-Qwen and its original base Qwen2.5-Math. As shown in Figure \ref{fig:sub-response-length}, on the GSM8k and MATH datasets, although the initial output length of Deepseek-R1-Distill-Qwen is greater than that of Qwen2.5-Math, the outputs of both models converge to the same length. In Figure \ref{fig:sub-accuracy}, on the MATH dataset, the accuracy of Deepseek-R1-Distill-Qwen is lower than that of Qwen2.5-Math due to its earlier output truncation. However, after fine-tuning with \ours, the accuracies of both Deepseek-R1-Distill-Qwen- and Qwen2.5-Math exhibit a steady increase.



We also analyze the qualitative differences. As shown in Table~\ref{tab:distill-vs-base}, the distilled model not only corrects the conceptual error in percentage conversion but also delivers a clearer and more structured explanation of the underlying computation. While the correction itself may reflect improvements in factual accuracy, the cleaner numerical logic and removal of redundant commentary suggest that structural refinement plays a supporting role. Thus, we believe this example offers evidence that performance gains under length-penalty and distillation can arise from improved reasoning organization, not just from truncation alone. 

\begin{table*}[h]
\centering
\renewcommand{\arraystretch}{1.3}
\begin{tabular}{|p{4.2cm}|p{5.7cm}|p{5.7cm}|}
\hline
\textbf{Dataset / Question} & \textbf{Base Model Output (Qwen2.5-Math-7B)} & \textbf{Distilled Model Output (Deepseek-R1-Distill-Qwen-7B)} \\
\hline
\textit{What is the positive difference between \(120\%\) of 30 and \(130\%\) of 20?}
& Incorrect due to misinterpreting percentages: \newline
\textit{``120\% of 30 is \(0.12 \times 30 = 3.6\). 130\% of 20 is \(0.13 \times 20 = 2.6\). The positive difference is \(3.6 - 2.6 = 1\). \(\boxed{1}\)''}
& Correct and concise: \newline
\textit{``120\% of 30 is \(1.2 \times 30 = 36\). 130\% of 20 is \(1.3 \times 20 = 26\). The difference is \(36 - 26 = \boxed{10}\).''}  \\
\hline
\end{tabular}
\caption{Comparison of reasoning behavior between the base and distilled models after applying \ours. The base model misinterprets percentage values due to incorrect decimal logic and lacks structural clarity. The distilled model applies correct numeric conversions, removes redundant phrasing, and presents a concise, high-confidence reasoning chain—illustrating how compression improves both accuracy and readability.}
\label{tab:distill-vs-base}
\end{table*}

\end{document}